\documentclass[letterpaper, 10 pt, conference]{ieeeconf}  % Comment this line out if you need a4paper

\IEEEoverridecommandlockouts                              % This command is only needed if 
\usepackage{graphics} % for pdf, bitmapped graphics files
\usepackage{epsfig} % for postscript graphics files
\usepackage{times} % assumes new font selection scheme installed
\usepackage{amssymb}  % assumes amsmath package installed
\usepackage{amsmath} % assumes amsmath package installed
\usepackage{subfigure}
\usepackage{stfloats}
\usepackage{makecell}
\usepackage{mathrsfs}
\usepackage{cite}
\usepackage{multirow}
\usepackage{booktabs}

\title{\LARGE \bf
Agile Formation Control of Drone Flocking Enhanced with Active Vision-based Relative Localization
}

\author{Peihan Zhang, Gang Chen, Yuzhu Li and Wei Dong$^*$% <-this % stops a space
\thanks{This work is partially supported by the Scientific and Technical Innovation 2030-``Artificial Intelligence of New Generation'' Major Project (2018AAA0102704) and National Natural Science Foundation of China (Grant No. 51975348, 51605282). The imaging processing section is also partially supported by grants from NVIDIA cooperation.\textit{(Corresponding author: Wei Dong.)}}% <-this % stops a space
\thanks{The authors are with the State Key Laboratory of Mechanical System and Vibration, School of Mechanical Engineering, Shanghai Jiao Tong University, Shanghai 200240, China {(email: xiancaiguazi@sjtu.edu.cn; chg947089399@sjtu.edu.cn; yuzhu\_0222@sjtu.edu.cn; dr.dongwei@sjtu.edu.cn)}.}%
}

\begin{document}

\maketitle

\thispagestyle{empty}
\pagestyle{empty}

%%%%%%%%%%%%%%%%%%%%%%%%%%%%%%%%%%%%%%%%%%%%%%%%%%%%%%%%%%%%%%%%%%%%%%%%%%%%%%%%
\begin{abstract}
The vision-based relative localization can provide effective feedback for the cooperation of aerial swarm and has been widely investigated in previous works. However, the limited field of view (FOV) inherently restricts its performance. To cope with this issue, this letter proposes a novel distributed active vision-based relative localization framework and apply it to formation control in aerial swarms. Inspired by bird flocks in nature, we devise graph-based attention planning (GAP) to improve the observation quality of the active vision in the swarm. Then active detection results are fused with onboard measurements from Ultra-WideBand (UWB) and visual-inertial odometry (VIO) to obtain real-time relative positions, which further improve the formation control performance of the swarm. Simulations and experiments demonstrate that the proposed active vision system outperforms the fixed vision system in terms of estimation and formation accuracy. 
%  The swarm with active vision achieves agile flocking movements with an acceleration of 4 $\boldsymbol{m/s^2}$ in circular formation tasks and a estimation accuracy. 

\end{abstract}

\begin{keywords}
Aerial swarm, relative localization, formation control, active vision. 
\end{keywords}

%%%%%%%%%%%%%%%%%%%%%%%%%%%%%%%%%%%%%%%%%%%%%%%%%%%%%%%%%%%%%%%%%%%%%%%%%%%%%%%%
\section{Introduction}
Aerial swarms have gained an increasing research focus in recent years, owing to their promising applications in cooperative missions, such as exploration, inspection, search and rescue \cite{shakhatreh2019unmanned}. The swarms outperform an individual flying robot in terms of capability,  flexibility and survivability \cite{chung2018survey}. To fully realize collaboration in a swarm, relative localization is a fundamental part \cite{coppola2020survey}. Such localization provides a basis for collision avoidance, formation control and other swarm behaviors \cite{bouffanais2016design}. 

The scalability and independence on communication make vision an ideal candidate for relative localization of distributed aerial swarms \cite{schilling2021vision}. Therefore, visual sensors have been widely adopted to obtain relative positions in previous works \cite{dias2016board, walter2018fast, walter2019uvdar, xu2020decentralized, vrba2020marker, schilling2021vision, xu2021omni}.  However, the vision-based approach is inherently restricted by limited field of view (FOV). To overcome this limitation, it is intuitive to realize omnidirectional vision detection. Attempts in the literature include fisheye camera \cite{walter2018fast, xu2021omni} and camera array \cite{saska2016formations}. Nevertheless, the fisheye camera requires additional computational resources to rectify distortion and an array of visual sensors makes the system bulky.

\begin{figure}[htpb]
      \centering
      \subfigure[The aerial platform with active vision system.]
        {\includegraphics[scale=0.2]{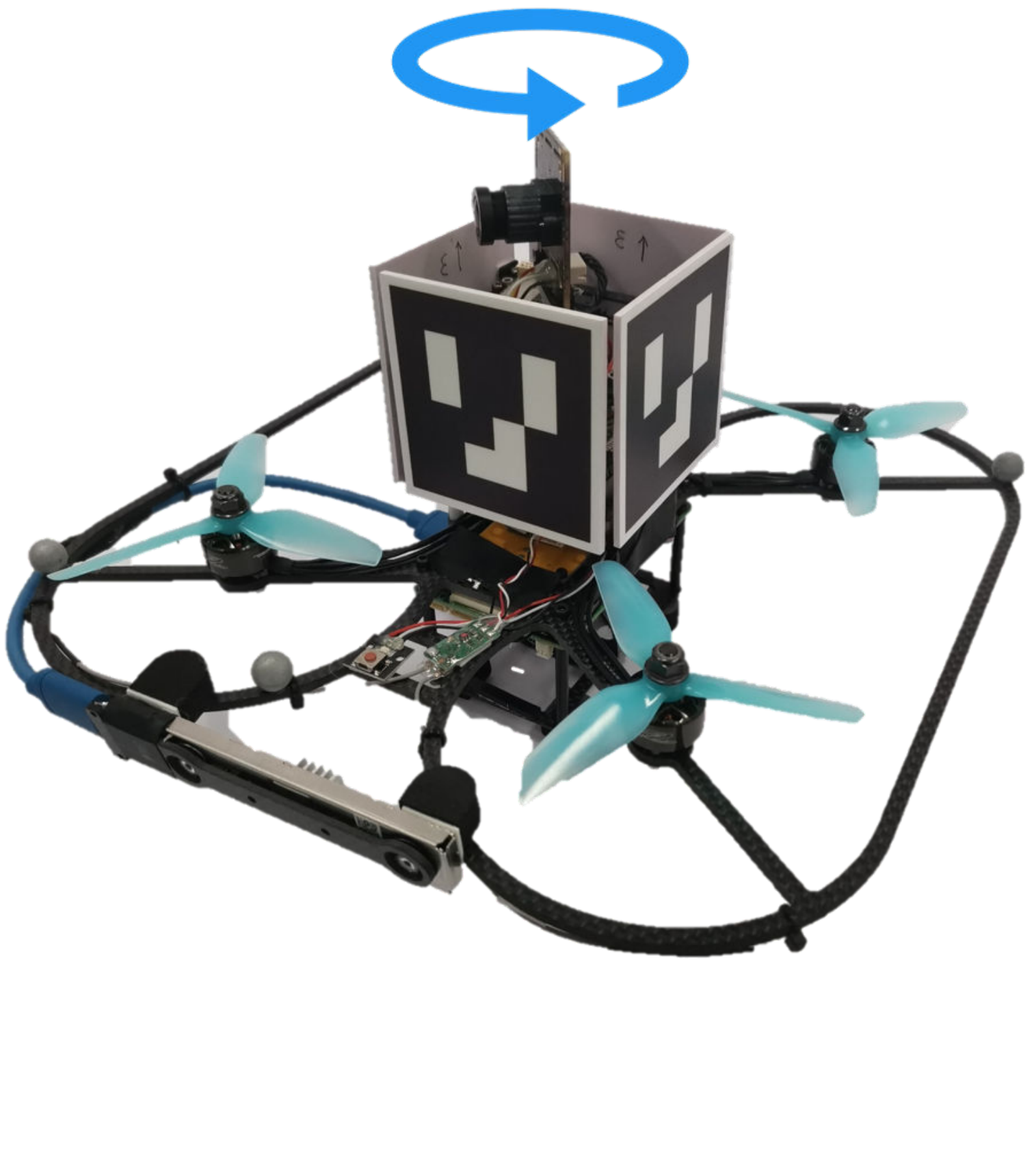}
        \label{fig:drone-structure}
        }
      \subfigure[Outdoor formation with 4 drones.]
        {\includegraphics[scale=0.2]{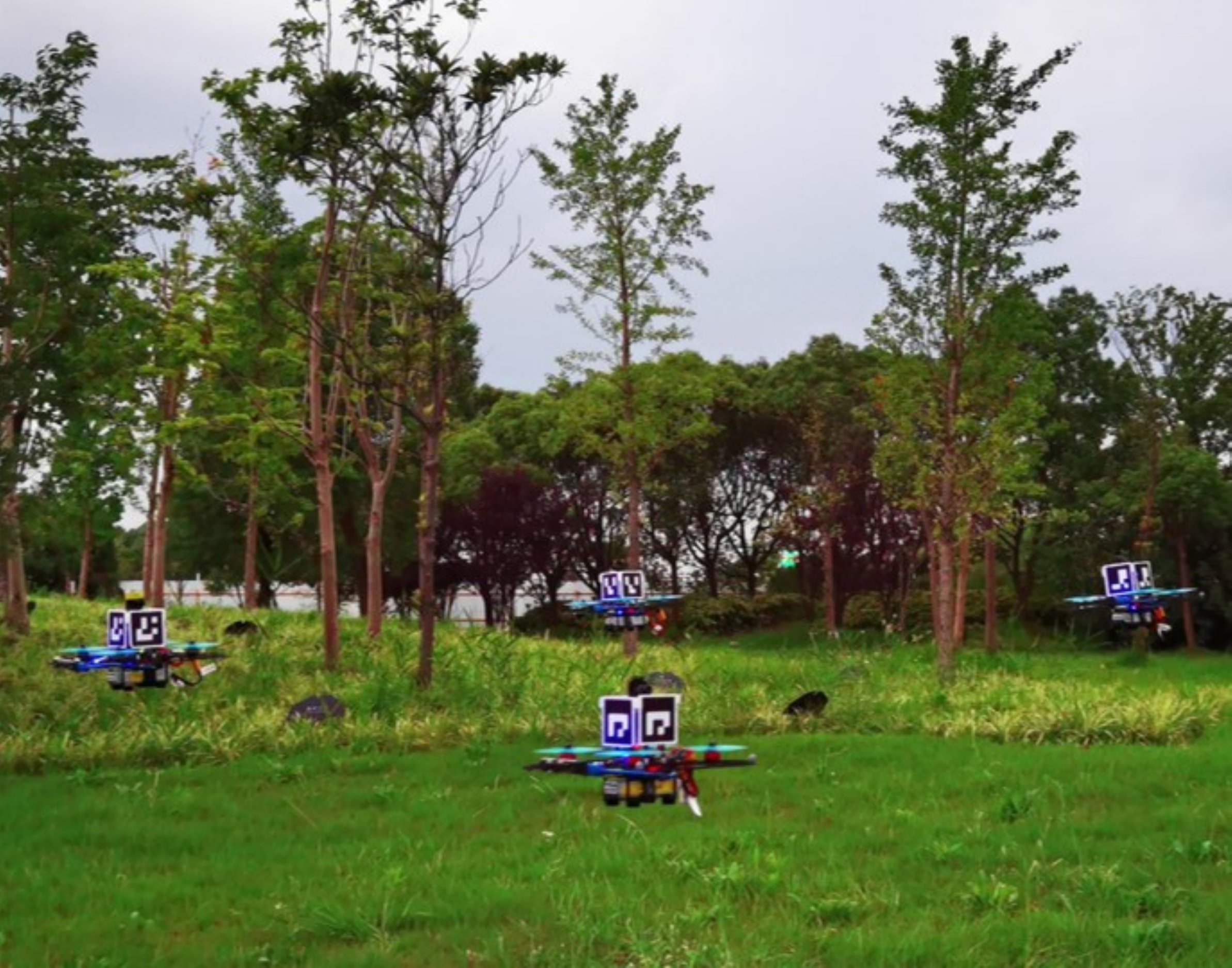}}
      \caption{Drone flocking enhanced with active vision. Each drone actively observes the other drones to improve formation accuracy.}
      \label{swarm-side1}
\end{figure}

% Owing to the similarity between bird flocks and unmanned aerial swarms, the mechanisms in bird flocks have reference significance for aerial swarms \cite{qiu2017multiple}. Vision is the primary sensory modality that enables relative localization in bird flocks \cite{nagy2010hierarchical}.
Flocks in nature shed light on the design of an efficient omnidirectional vision detection. Vision is a critical component for birds to respond to their neighbors' motion when flying in flocks \cite{vidal2004following, nagy2010hierarchical}. The eyeballs or body movement enable a bird to observe its surroundings better \cite{tyrrell2017avian}. 
% in bird flocks will turn its head around to observe others to maintain the formation and avoid collisions. 
In a similar notion, active vision leverages the physical motion of the camera to dynamically gather more information about surroundings \cite{chen2011active}. Compared with fixed vision, active vision overcomes the restriction of FOV without adding extra visual devices. Active vision has been successfully applied to aerial swarm missions, such as human detection of marine search and rescue \cite{qingqing2020towards} and target tracking with optimal view-point configurations \cite{tallamraju2019active}, outdoor motion capture for human pose estimation \cite{kiciroglu2020activemocap}. Applying active vision to aerial swarms requires decentralized planning of temporal and spatial distribution of the camera's attention, i.e., planning when and where the camera observes so that all drones can cooperate and achieve accurate and agile flight.

% There are at least two challenges for applying active vision to relative localization in aerial swarms. First of all, the vision-based approach itself is subject to blur \cite{coppola2020survey}, and the movement of active vision will aggravate the motion blur and affect the quality of observation. Benefiting from active motion of camera while being affected from blur as little as possible is a challenge for active vision. Second, although active vision can overcome the limited FOV through motion, the scope of the camera observation is still limited at given time. Therefore, planning the temporal and spatial distribution of the camera's observation focus, i.e., planning when and where the camera observes,  is still an open challenge. For swarm tasks, the question is more challenging to schedule the observation of all agents to maxmize obervation quality. For example, this planning can aim to minimize the uncertainty of observation \cite{tallamraju2019active}. 

To the best of our knowledge, there lacks an active vision-based approach to tackle restricted FOV for relative localization in aerial swarms. Therefore, in this letter, we propose a fully distributed active vision-based framework for real-time relative localization of aerial swarms. This framework is infrastructure-free, i.e., eliminating requirements for external devices such as global positioning system (GPS) and the motion capture system. An independent rotational degree of freedom (DOF) is introduced to the camera to achieve the active vision. We also take advantage of the fusion scheme, which utilizes measurements from Ultra-WideBand (UWB), visual-inertial odometry (VIO) and active vision detection to realize the robust omnidirectional relative estimation. The estimation is then applied to agile formation control tasks. 
% We devise a graph-based approach to evaluate the observation schedule of cameras in aerial swarms so that formation control can be realized accurately.

% Some works combine the methods mentioned above to reach better relative localization. In \cite{nguyen2019distance}, UWB and odometry sensors are combined without external positioning systems. In \cite{xu2020decentralized}, estimation from stereo cameras, VIO and UWB module are fused to implement an on-board real-time relative state estimation system in aerial swarm frameworks. Despite its high accuracy, \cite{xu2020decentralized} has difficulty initializing swarm positions based on pure visual detection due to out of FOV problem. \cite{xu2021omni} overcomes the problem of limited FOV by using fisheye cameras to realize omnidirectional detection. Nevertheless, the fisheye camera requires additional computational resources to deal with distortion. Moreover, the schemes mentioned above concentrate on relative localization while control-oriented frameworks are more desired in multi-robot formation control tasks.

The main contributions of this article are as follows: 1) A novel active vision-based relative localization framework, which overcomes limited FOV and achieves centimeter-level accuracy. 2) A graph-based attention planning (GAP) algorithm, which is coupled with the swarm formation and provides optimal active observation planning for the swarm.

\section{Related Works}
Relative localization is a prerequisite for the cooperation of aerial swarms \cite{walter2019uvdar}. External devices, such as GPS, motion capture system and UWB positioning system with anchors, are adopted to obtain absolute positions and deduce the relative positions between agents. The dependence on the external infrastructure of this approach restricts the deployment of the system to unknown environments \cite{ziegler2021distributed}. Also, it is based on a centralized framework, so it cannot meet the requirements in fully distributed control of swarms \cite{xu2021omni}.

To tackle this issue, methods relying on onboard devices in the distributed framework are proposed. 
% These approaches can be divided into three categories: ego-localization-based approach, distance-based approach and vision-based approach. 
A straightforward way is relying on ego-state estimations.
This approach utilizes onboard local ego-state estimation to obtain the relative localization in the common reference frame. This estimation comes from onboard computation, such as VIO \cite{weinstein2018visual}, simultaneous localization and mapping (SLAM) \cite{lajoie2020door}. This approach requires a known initial position of each agent due to indirect measurement, and suffers from the drift issue \cite{xu2020decentralized}, potentially leading to formation failure or even collisions \cite{li2020autonomous}. Moreover, the requirement to share ego-positions between robots makes the system unfeasible to cope with highly dynamic environments with high communication throughput \cite{dias2016board}.

Another solution is to rely on distance sensors. Distance sensors have the ability to measure distance between individuals in an omnidirectional way directly. UWB, one popular distance sensor, has been adopted widely in the literature. External UWB position system based on fixed anchors can estimate positions of agents in the common reference frame \cite{ziegler2021distributed}, but the requirement of the previous deployment of anchors means the framework can hardly be adopted in the unknown environment\cite{xu2020decentralized}. Efforts have been made to design an anchor-free UWB position scheme, such as UWB-IMU coupled approach \cite{cossette2021relative,shalaby2021relative}. However, 
% due to the single-dimensional distance measurement, 
the accuracy of relative localization relying barely on UWB is not satisfactory \cite{cossette2021relative,shalaby2021relative}. To achieve better estimation, UWB measurements are fused together with measurements from other sensors, such as VIO \cite{ziegler2021distributed}, wheel encoder \cite{li2020relative}, and optical flow \cite{guo2019ultra, nguyen2019distance}.

Unlike distance sensors, visual sensors can provide 3-D position estimations of targets. The vision-based approach detects the relative positions of other agents using onboard cameras. The detection can be achieved by pre-known markers attached to agents \cite{walter2018fast, walter2019uvdar}, or by a pre-trained convolutional neural network (CNN) detection algorithm \cite{xu2020decentralized, vrba2020marker}. Despite the scalability of the vision-based approach \cite{schilling2021vision}, estimation accuracy will deteriorate in the non-line-of-sight case due to restricted FOV \cite{xu2021omni}. To overcome this limitation, attempts have been made to enlarge the FOV of visual detection in the literature. In \cite{xu2021omni}, distortion-free images extracted from the fisheye camera are used to detect other drones. In \cite{schilling2021vision}, four cameras are installed to provide omnidirectional visual inputs. Nevertheless, addressing limited FOV with the fisheye camera or camera array is at the cost of additional mass, size, and computing power, which in turn brings new burdens \cite{coppola2020survey}. There is still room for improvement in omnidirectional vision detection to overcome FOV restriction.

\section{System Overview and Problem Formulation}
\subsection{System Overview}
Active vision aims to improve the detection performance of cameras limited by FOV and provide reliable and abundant position data of other agents for the estimator. To overcome the FOV limitation, we design a flexible structure to achieve the active vision as shown in Fig.\ref{fig:drone-structure}. This structure consists of a servo motor and a camera. They are mounted on a quadrotor's upper platform so that the motor can drive the camera in all directions to observe environments. Although active vision can also be implemented by moving the yaw angle of the drone, our approach benefits the control of the quadrotor. Due to the lightweight of the camera compared with the body of the quadrotor, this implementation brings negligible influence to the control of the original system while providing a flexible view field \cite{chen2021active}.

\subsection{Problem Formulation}
For an aerial swarm system that contains $N$ drones, the visual observations between drones can be represented by graph $\mathcal{G}_o = (\mathcal{V}, \mathcal{E})$, where the vertice set $\mathcal{V}$ represents $N$ drones and an edge $\boldsymbol{e}_i=\{v_j, v_k\}$ in the edge set $\mathcal{E}$ represents that drone $N_k$ is observed by $N_j$. We introduce $M \times N$ incidence matrix $\mathcal{D}(\mathcal{G}_o)$ to represent the visual connection, which is defined as

\begin{small}
\begin{equation}
\mathcal{D}\left(\mathcal{G}_o\right)=\left[d_{i j}\right], \text {where } d_{i j}=\left\{\begin{array}{cl}
-1 &\text {$v_j$ is the tail of $\boldsymbol{e}_i$}\\
1 &\text {$v_j$ is the head of $\boldsymbol{e}_i$}\\
0 &\text {otherwise}
\end{array}\right.
\end{equation}
\end{small}

% Let $N$ edges in $\mathcal{E}$ correspond to the order of $N$ vertices, representing each visual observation from the corresponding agent. This active vision observation determines the connection of the graph $\mathcal{G}_o$. We introduce $N \times N$ incidence matrix $\boldsymbol{D}(\mathcal{G}_o)$ to represent this connection, which is defined as

% \begin{small}
% \begin{equation}
% \boldsymbol{D}\left(\mathcal{G}_o\right)=\left[d_{i j}\right], \text { where } d_{i j}=\left\{\begin{array}{l}
% -1, \{v_i, v_j\} \in \mathcal{E} \\
% 1, i=j \\
% 0, \text { otherwise }
% \end{array}\right..
% \end{equation}
% \end{small}

Then the problem of optimal observation of active vision can be described as: 
\begin{equation}
\mathop{\arg\min}\limits_{\mathcal{D}(\mathcal{G}_o)}
{\Psi} (\mathcal{V}, \mathcal{E}).
\end{equation}

${\Psi} (\mathcal{V}, \mathcal{E})$ is an evaluation function to determine the quality of visual observations of the swarm.

% For an arbitrary drone $i$ of the swarm, determine which neighbor it should observe at time $t$ to facilitate the swarm's relative localization.

\section{Method}

In this section, we first discuss a solution to the problem mentioned above by providing graph-based attention planning. Later, the optimization-based relative localization fusing three kinds of measurements is explained. After that, the initialization of swarm positions by the estimator is addressed. Finally, the design of the formation control law is stressed, utilizing the results of relative localization.

\subsection{Graph-based Attention Planning}
Building on the graph $\mathcal{D}(\mathcal{G}_o)$ mentioned above, we introduce graph-based attention planning, which aims to find an active observation plan with minimal cost. In order to evaluate the attention of the active vision and improve the visual detection quality within the swarm, two main factors are considered: the observation distance and the flight direction of the drone. First, according to the feature of visual detection, the measurement errors increase with distance from the observer \cite{schilling2021vision}. Hence, a smaller observation distance is preferred. Since the relative localization problem is integrated with the formation control task in our system, the planning computes observation distance according to the desired formation, which improves computing efficiency and makes offline pre-planning possible. Second, more observation in the flight direction of the drone can help preclude collisions with neighbors and thus should be concerned.

% agents in the aerial swarm are sparsely distributed in space, so an agent generally only observe one neighbor at a time in most cases. As a result, we assume there are $N$ edges in $\mathcal{E}$ corresponding to the order of $N$ vertices, representing each visual observation from corresponding agent. This active vision observation determines the connection of the graph $\mathcal{G}_o$. We introduce $n \times n$ incidence matrix $\boldsymbol{D}(\mathcal{G}_o)$ to represent this connection, which is defined as

Let $\boldsymbol{v}$ and $\boldsymbol{x}$ represent the velocities and positions of drones respectively in the global frame, i.e., $\boldsymbol{v}=\left[\begin{array}{llll}\boldsymbol{v_{1}}, \boldsymbol{v_{2}} & , \cdots & \boldsymbol{v_{N}}\end{array}\right]^{T}$, $\boldsymbol{x}=\left[\begin{array}{llll}\boldsymbol{x}_{1}, \boldsymbol{x}_{2} & , \cdots & \boldsymbol{x_N}\end{array}\right]^{T}$.
Let ${\mathcal{D}}_o(t)$ represent the incidence matrix $\mathcal{D}(\mathcal{G}_o)$ at time $t$, and graph Laplacian of $\mathcal{G}_o$ is $\mathcal{L}_o(t)={\mathcal{D}_o(t)}^T\mathcal{D}_o(t)$. Then the cost function of the GAP is described as
\begin{equation}
\label{eq:cost}
\begin{aligned}
 \min _{{\mathcal{D}_o(t)}}
\ & f=  \gamma_1 \left({\mathcal{D}}_o(t) \boldsymbol{x} \right)^{T}{\mathcal{D}}_o(t)\boldsymbol{x}-\gamma_2 \boldsymbol{v}^{T} {\mathcal{D}}_o(t) \boldsymbol{x}
\\  s.t.\quad & \mathcal{D}_o(t) \cdot \boldsymbol{1}_{N \times 1}=\boldsymbol{0}_{N \times 1}
\\            & \lambda_1(\mathcal{L}_o(t))=0, \lambda_k(\mathcal{L}_o(t))>0
\\            & \min_{1 \leq j \leq N} \sum_{i=1}^{M}\vert{d_{ij}}\vert \geq 2
\end{aligned}
\end{equation}
where $k=2,3,...,N$ and $\gamma_1$ and $\gamma_2$ are positive weights. The first term in the cost function \eqref{eq:cost} is the sum of observation distances in the swarm at time $t$. A smaller sum of distances is desired, making a drone prefer to observe another near drone in the formation and thus improving visual detection quality. The second term means the inner product of each pair of velocity and observation direction. According to the nature of the inner product, the more consistent the direction of velocity and the observation, the smaller the cost function is.

To find the optimal solution, we make two simplifying assumptions according to our scenario:
\begin{itemize}
 \item[1)] $\mathcal{D}_o(t)^T \cdot \boldsymbol{1}_{M \times 1}=\boldsymbol{0}_{M \times 1}$
 \item[2)] $\min_{1 \leq j \leq N} \sum_{i=1}^{M}\vert{d_{ij}}\vert = 2$
\end{itemize}
% \begin{itemize}
%  \item[1)] There are $N$ edges in $\mathcal{E}$. An agent generally only observe one neighbor at a time in most cases because agents in the aerial swarm are sparsely distributed in space.
%  \item[2)] The agent's position difference in the vertical direction is negligible. 
%  \item[3)] The results of active vision detection are shared among the swarm, so each agent is able to obtain the information of observation within the swarm.
% \end{itemize}

\begin{figure}[thpb]
    \centering
    \includegraphics[scale=0.65]{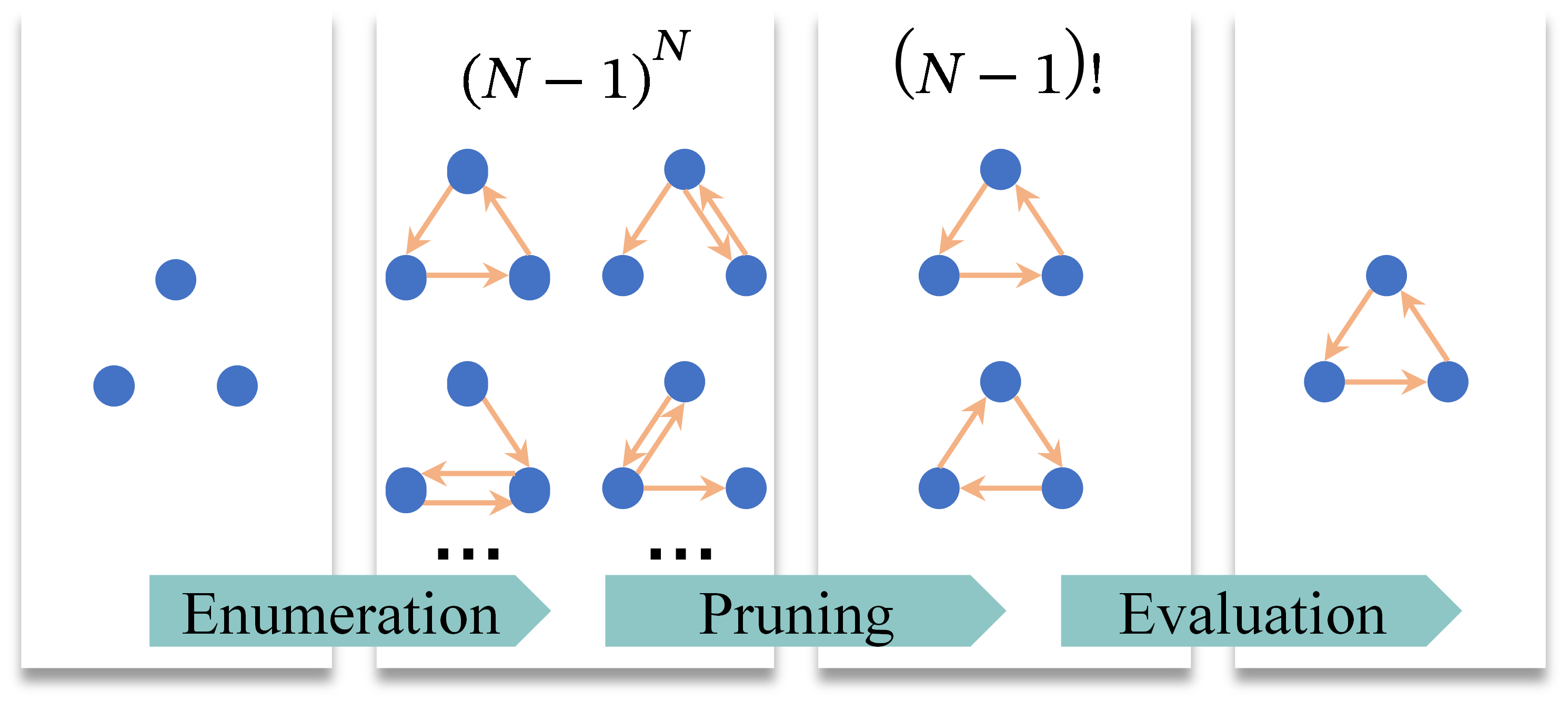}
    \caption{The flow of GAP.}
    \label{fig:GAP-flow}
\end{figure}

The first assumption means a drone generally could observe one neighbor at a time in most cases because drones in the aerial swarm are sparsely distributed in space. As a result, the target will be placed at the center of the camera's FOV for better tracking and observation. The second assumption means each drone will be observed so that the detection results could be shared among the swarm for distributed estimation. Based on our assumptions, the possible number of connections is reduced from $(N-1)^N$ to $(N-1)!$. Take the formation of 4 drones in our experiments as an example. There are 81 and 6 possible connections before and after pruning, respectively. Searching for the optimal solution of GAP becomes acceptable after pruning. The flow of GAP for 3 drone connection planning is shown in Fig.\ref{fig:GAP-flow}. 

% To find the optimal solution, possible connections of $\mathcal{G}_o(t)$ are evaluated. Considering that the formation in our scenario is predefined, the solution can be obtained offline in advance. Based on our assumptions, an agent observes only one neighbor, so each agent has $N-1$ possible observations. For a swarm of $N$ agents, there are $(N-1)^N$ connections, which is a large space to evaluate with the growth of agent number. Considering computational tractability, we keep connections that provide abundant visual information while pruning others. Abundant information refers to the case where each agent is assigned an observation target and is observed simultaneously. In other words, there exists a loop in the graph $\mathcal{G}_o(t)$. If so, the swarm will have sufficient visual detection results of every agent, facilitating the distributed relative position estimator.  After the pruning, there are $(N-1)!$ connections. Take the formation of 4 agents in our experiments as an example. There are 81 and 6 possible connections before and after pruning, respectively. Searching for the optimal solution by enumeration becomes acceptable after pruning. The flow of GAP is shown in Fig.\ref{fig:GAP-flow}. 

After the assignment of the detection target by GAP, the rotation angle of the camera can be determined according to the relative position of the observation target in the drone's body frame. Since drones fly at nearly the same altitude and the camera only has one rotational DOF along the Z-axis, the relative position is calculated based on the horizontal plane. 
% As mentioned above, each agent will observe one neighbor in most cases, so the target will be placed at the center of the camera's FOV for better tracking and observation.

% This optimal solution represents a connected graph and the orientation of the edge determines the direction of the camera rotation angle. Suppose there is a connection from agent $i$ to agent $j$ in the optimal solution and the relative position between them is $\boldsymbol{X_{i j}}$, then the camera rotation angle $\theta_{i}$ of agent $i$ can be represented as
% \begin{equation}
%  \theta_{i} = atan2\left(\boldsymbol{X_{i j}}(2), \boldsymbol{X_{i j}}(1)  \right)
% \end{equation}
% 
% Since the camera has only one rotational DOF along the Z-axis, we mainly concern about the directions in the $xy$ plane. The $\boldsymbol{X_{i j}}(2)$ and $\boldsymbol{X_{i j}}(1)$ means the relative position value of Y-axis and X-axis respectively. As mentioned above, each agent will observe one neighbor in most cases, so the choice of $\theta_{i}$ is to place the observed individual in the center of the agent's FOV for better tracking and observation.

\subsection{Relative Localization}

\begin{figure}[thpb]
    \centering
    \includegraphics[scale=0.3]{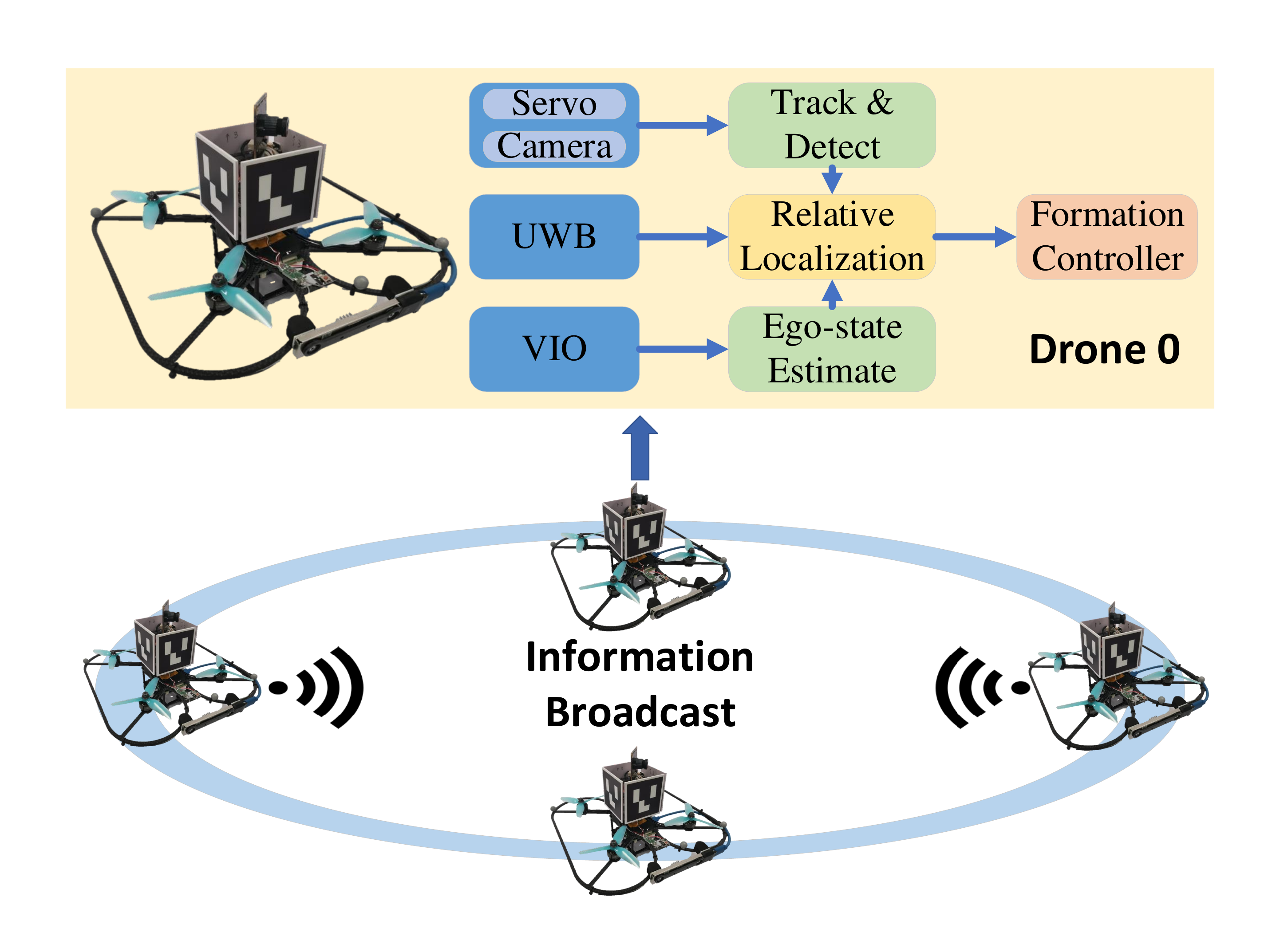}
    \caption{The framework of active vision-based relative localization. The measurements from active vision, VIO and UWB are fused to obtain robust and real-time estimations. Then these estimations are integrated with formation task.}
    \label{fig:system-framework}
\end{figure}

Although active vision detection can provide relative positions, some drones may be invisible due to occlusion or beyond visual range. Moreover, estimations from vision can be intermittent due to misdetection \cite{falanga2017aggressive}. Hence, active vision measurements could be complemented by other sensors so that we can obtain robust and continuous estimations. Inspired by the efficacy of the fusion scheme in \cite{xu2020decentralized}, measurements from UWB and VIO are also adopted. In this subsection, we first discuss the details of active vision and the other two measurements. Then we introduce the implementation of optimization-based relative localization. The framework of our approach is shown in Fig.\ref{fig:system-framework}.

\subsubsection{Active Vision Measurement}
The active vision system of the drone includes a servo motor that drives the camera in all directions to observe environments. In order to detect and identify other drones, ArUco markers are adopted. To realize omnidirectional detection, the placement of markers should be visible from any direction. Considering the constraints of space of an aerial platform and the fact that these platforms are often deployed at similar altitudes, we attach four markers around each aerial platform as Fig.\ref{drone-device} shows. The four markers have the same ID number and each aerial platform has a unique ID number so that aerial platforms can be identified directly through marker ID. Compared to CNN-based approach, our approach does not require pre-training and labeling. Also, real-time detection can be realized without GPU. These benefits make the deployment of our active vision system convenient.

\subsubsection{Visual-Inertial Measurement}
The data from the commercial Intel RealSense T265 tracking module is regarded as visual-inertial odometry measurement. Basically, the VIO serves as an ego-state estimator for the flight controller and formation controller. Because the VIO provides the position of a drone in the local frame, knowing the swarm's initial positions is the prerequisite to utilize the VIO measurement. The initialization is discussed in \ref{sec:initial}. After that, the relative positions can be deduced from VIO displacement based on initial positions.

\subsubsection{UWB Measurement}
Each drone in the aerial swarm is equipped with a UWB module. Direct inter-agent distances can be obtained from each pair of UWB nodes. Because UWB measurement may yield significant outliers due to interferences, the Savitzky Golay filter is adopted to process UWB measurements.

% \subsubsection{Optimization-based Relative Localization}
To fuse the measurements from active vision detection, VIO and UWB, an optimization-based algorithm is adopted. For drone $N_i$ in the aerial swarm, $3\times(N-1)$ variables need to be estimated at time frame $k$. These variables represent the relative positions in three-dimensional space between drone $N_i$ and the other $N-1$ drones in drone $N_i$'s local frame. The optimization function is expressed as the following formulation:

\begin{equation}
\begin{aligned}
\min _{\boldsymbol{\hat{x}}_{i j}^{k}}
\ J
=&\sum \left\|\boldsymbol{p}_{i j}^{k}- \boldsymbol{\hat{x}}_{i j}^{k}\right\|_{2}+\sum \left\|\boldsymbol{p}_{j i}^{k}+\boldsymbol{\hat{x}}_{i j}^{k}\right\|_{2} + \\
&\sum \left|\left\|\boldsymbol{\hat{x}}_{i j}^{k}\right\|_{2}-d_{i j}^{k}\right| +\\
&\sum \left\|\boldsymbol{x}_{VIO,j}^{k}-\boldsymbol{x}_{VIO,i}^{k}-\boldsymbol{\hat{x}}_{i j}^{k}\right\|_{2} +\\
&\sum \left\|\boldsymbol{\hat{x}}_{ij}^{k-1}+\left( \boldsymbol{v}_{j}^{k}-\boldsymbol{v}_{i}^{k} \right)\delta t -\boldsymbol{\hat{x}}_{i j}^{k}\right\|_{2}
\end{aligned}
\label{eq:J}
\end{equation}

where $\boldsymbol{\hat{x}}_{i j}^{k}$ refers to an estimation of the relative position between drone $N_i$ and drone $N_j$ at time frame $k$. The first term is composed of two possible residuals of active vision detection measurements; $\boldsymbol{p}_{ij}^{k}$ means the pair of valid detection of drone $N_j$ detected by drone $N_i$; $\boldsymbol{p}_{j i}^{k}$ means the pair of valid detection of drone $N_i$ detected by drone $N_j$, obtaining through communication; $\left|\left\|\boldsymbol{\hat{x}}_{i j}^{k}\right\|_{2}-d_{i j}^{k}\right|$ represents the residual of UWB distance measurements; $\boldsymbol{x}_{VIO,j}^{k}$ and $\boldsymbol{x}_{VIO,i}^{k}$ represents the VIO measurements from drone $N_j$ and drone $N_i$ at time frame $k$ in global frame respectively. To leverage the dynamic of the system, the first order expectation is introduced in the last term; $\boldsymbol{v}_{j}^{k}-\boldsymbol{v}_{i}^{k}$ represents relative velocity between drone $N_j$ and drone $N_i$; $\delta t$ represents time interval between time frame $k-1$ and $k$; $\boldsymbol{\hat{x}}_{ij}^{k-1}+\left( \boldsymbol{v}_{j}^{k}-\boldsymbol{v}_{i}^{k} \right)\delta t$ represents expected relative position at time frame $k$ based on time frame $k-1$.

The estimation is implemented in a distributed manner. Each drone runs its estimator by leveraging information it collects or interchanges with other drones. Due to the asynchronous communication of UWB modules in the aerial swarm, drone $N_i$ may not obtain information of all other drones at time frame $k$. In this case, specific terms in the \eqref{eq:J} will be omitted due to incomplete data. We adopt the Ceres-solver to solve this non-linear least-squares optimization problem. Considering the movements of drones are continuous, we set the initial values of the solver as values of the last time frame $k-1$, which brings the benefit of the faster convergence of the solver.

\subsection{Initialization of Relative Positions}
\label{sec:initial}
Since the VIO measurement is in a drone's local frame, it is required to know drones' initial positions to determine their relative positions. The initialization of relative positions is implemented by fusing the results of active detection and UWB measurement. For convenience, the drones in the aerial swarm face the same directions by aligning their compasses and there is no rotation between the local frame and global frame during the initialization stage. The framework of initialization is similar to that of the above section, which is described as:

\begin{equation}
\begin{aligned}
\min_{\boldsymbol{\hat{x}}_{i j}^{0}}
% \ J=& J_{ArUco}+J_{UWB} \\
J=& \sum_{k \in T}\left\|\boldsymbol{p}_{i j}^{k}-\boldsymbol{\hat{x}}_{i j}^{0}\right\|_{2}+\sum_{k \in T}\left\|\boldsymbol{p}_{j i}^{k}+\boldsymbol{\hat{x}}_{i j}^{0}\right\|_{2} +\\
&\sum_{k \in T}\left|\left\|\boldsymbol{\hat{x}}_{i j}^{0}\right\|_{2}-d_{i j}^{k}\right|
\label{eq:Jinit}
\end{aligned}
\end{equation}

where $\boldsymbol{\hat{x}}_{i j}^{0}$ represents the estimated initial relative position between drone $N_i$ and drone $N_j$. $T$ represents the period when the initialization program collects data. Because all drones are static, their relative positions are regarded as constant and thus time-invariant. The initialization algorithm will utilize measurements of active detection and UWB within a short period. After initialization, the optimization algorithm will have access to VIO measurements to estimate relative localization.

\subsection{Formation Control}

In this subsection, we aim to apply the proposed relative localization framework to a consensus-based formation control task. We first consider the second-order system of the drone. Then we propose a formation control law.

In a multi-robot system, the outer-loop dynamics of drone $N_i$ can be approximately described by

\begin{equation}
 \left\{\begin{array}{l}
\boldsymbol{\dot{x}}_{i}(t)=\boldsymbol{v}_{i}(t) \\
\boldsymbol{\dot{v}}_{i}(t)=\boldsymbol{u}_{i}(t)
\end{array}\right.
\label{eq:dyn}
\end{equation}
where $\boldsymbol{x}_{i}(t) \in \mathbb{R}^{3}$, $\boldsymbol{v}_{i}(t)\in \mathbb{R}^{3}$ and $\boldsymbol{u}_{i}(t)\in \mathbb{R}^{3}$ denote the position, velocity and control input vectors respectively.

The formation controller adopts a forward feedback scheme. Let ${\boldsymbol{r}(t)}=\left[\begin{array}{llll}\boldsymbol{x_{1}}, \boldsymbol{x_{2}} & , \cdots & \boldsymbol{x_{N}}\end{array}\right]^{T}$, then the control law of the swarm is described by:
\begin{equation}
\begin{aligned}
% \boldsymbol{u}_{i}(t) = & \boldsymbol{\dot{v}}_{i}^{*}(t)+ 
%   K_{1}\left(\boldsymbol{x}_{i}^{*}(t)-\boldsymbol{x}_{i}(t)\right)+ 
%   K_{2}\left(\boldsymbol{v}_{i}^{*}(t)-\boldsymbol{v}_{i}(t)\right)+ \\ 
% &  K_{3} \sum_{j \in N_{i}} w_{i j}\left(\left(\boldsymbol{x}_{j}^{*}(t)-\boldsymbol{x}_{i}^{*}(t)\right)-\left(\boldsymbol{x}_{j}(t)-\boldsymbol{x}_{i}(t)\right)\right) \\
\boldsymbol{u}(t) = & \boldsymbol{\ddot{r}}^{*}(t)+ \phi(\boldsymbol{\dot{r}}^{*}(t), \boldsymbol{\dot{r}}(t), \boldsymbol{{r}}^{*}(t), \boldsymbol{{r}}(t))+\mathcal{L}_o(t)\boldsymbol{{r}}(t)
\end{aligned}
\end{equation}
where $\boldsymbol{\ddot{r}}^{*}(t)$, $\boldsymbol{\dot{r}}^{*}(t)$ and $\boldsymbol{{r}}^{*}(t)$ are expected acceleration, velocity and position of the swarm respectively. The second term denotes the control value caused by the position and velocity errors; the last denotes the control value caused by swarm formation. The last term utilizes the results of relative localization and ensures the formation accuracy of the aerial swarm.

\section{Experiment}

\subsection{Experiment Setup}
A distributed aerial swarm consisting of four aerial platforms is designed to verify the relative localization framework and the performance of the formation controller. Both hardware and software are developed for the experiments.

\begin{figure}[htbp]
      \centering
      \includegraphics[scale=0.4]{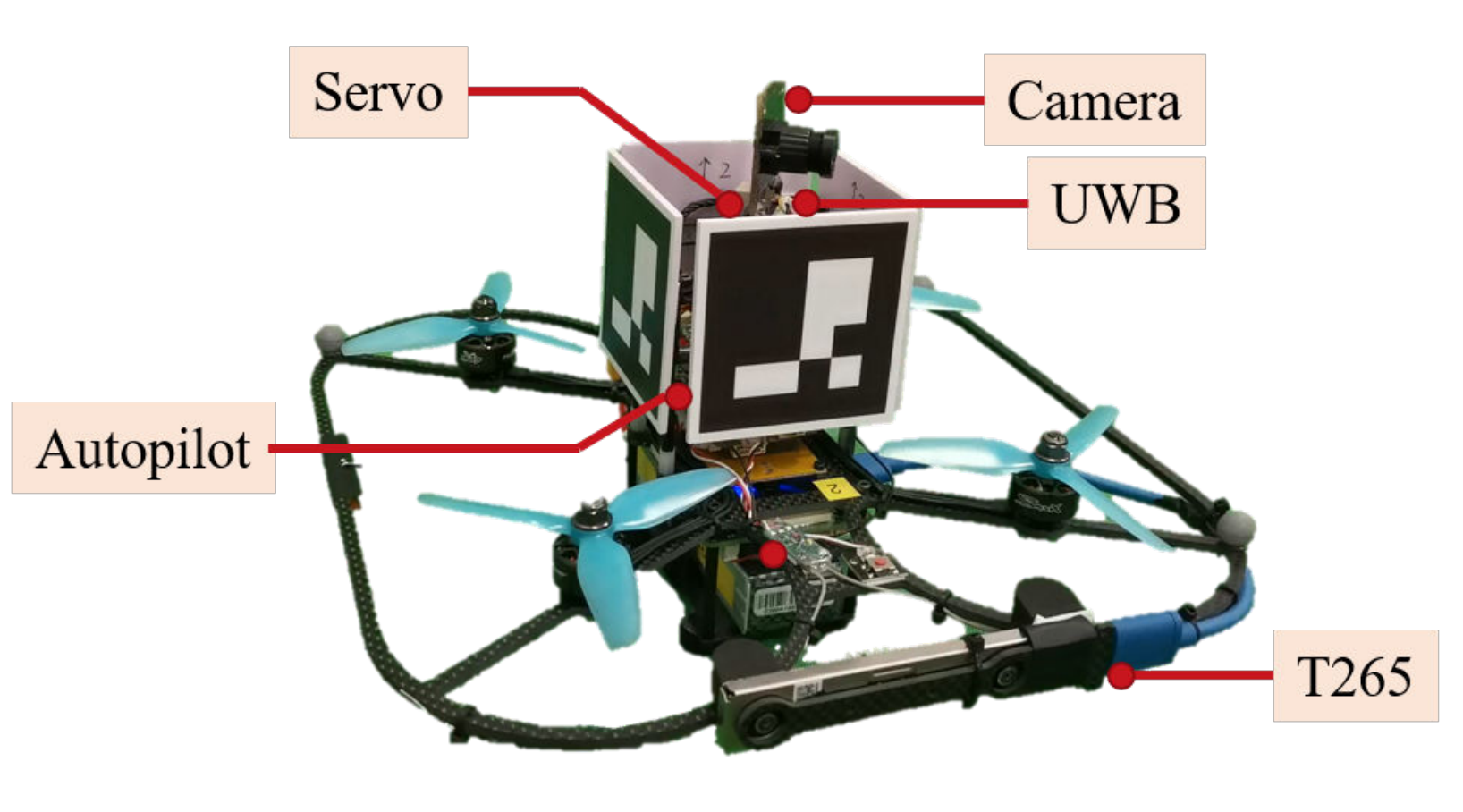}
      \caption{The aerial platform utilized in our experiments. This platform is equipped with a PixRacer flight controller, a monochrome camera, a servo motor, an Intel RealSense T265 tracking module, a Nooploop UWB module and a Up Core Plus on-board computer. Four ArUco markers are attached around the platform for identification.}
      \label{drone-device}
\end{figure}

We design a quad-rotor drone as the aerial platform (Fig.\ref{drone-device}). This platform is equipped with a PixRacer flight controller running PX4 firmware. A camera and a servo motor are combined to achieve active vision detection. The camera is a monochrome camera with a horizontal view field of 150$^{\circ}$ and 800$\times$600 resolution. The servo motor has an encoder providing 300 degrees rotation range. An Intel RealSense T265 tracking camera module is used for ego-state estimation. A UP Core plus computing board running ROS was adopted as the onboard computer. The mounted CPU was Intel Atom x7 (four cores, 1.8 GHz). A Nooploop UWB module is adopted for both inter-agent distance measurement and communication. The UWB module provides up to 25 Hz broadcasting frequency. Besides, four ArUco markers are attached to the drone to facilitate the identification.

The onboard computer runs Ubuntu 18.04, and all algorithms running on it are built under the ROS framework. We rotate the camera to capture the images and leverage OpenCV ArUco library to detect markers. The position data of markers is transformed to the local frame according to the drone's current pose and the current angle of the servo. The position data from the T265 tracking module, which is regarded as VIO measurement, is first transformed to the local frame and then is fused with visual detection measurements and UWB distance measurements for relative localization. The flight controller performs the basic attitude control while other computations are performed on the onboard computer.

The experiments are conducted in both indoor and outdoor scenarios. In indoor environments, the results of the VICON motion capture system are regarded as ground truth to compare with data from relative localization. In outdoor experiments, we only verify the swarm's ability to maintain formation without comparing with ground truth due to the lack of an external positioning system as the reference. In both indoor and outdoor environments, the aerial swarm utilizes the results of relative localization and performs formation control tasks.

\subsection{Experiment Result}
\label{subsec:result}
We first present the optimal observation planning obtained by GAP in the simulation. Then we report the results of extensive indoor experiments with the swarm of 4 drones and compare the performance of the proposed active vision with fixed vision in terms of relative localization accuracy, formation accuracy at different velocities. 

\begin{figure}[htp]
      \centering
    \subfigure[]
    {\includegraphics[scale=0.9]{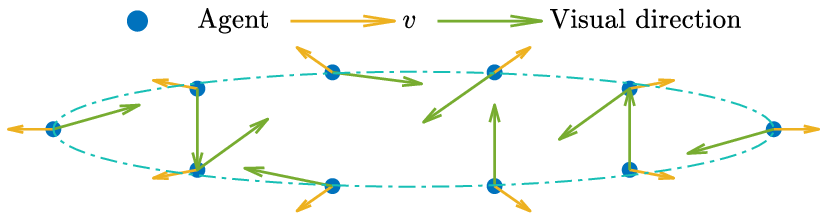}}
    \subfigure[]
    {\includegraphics[scale=0.9]{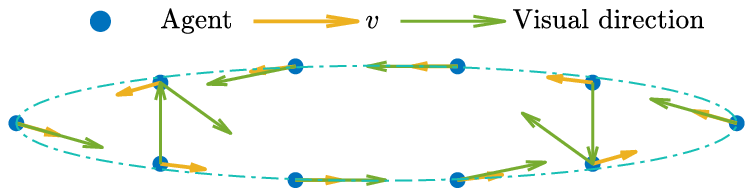}}
      \caption{The optimal connection of 10 drones obtained by GAP simulation at different velocity settings.} 
      \label{fig:optimal-connection}
\end{figure}

First of all, we design a simulation program in MATLAB to utilize graph-based attention planning to obtain the optimal observation offline. Simulation examples with 10 drones at two different velocity settings are shown in Fig.\ref{fig:optimal-connection}. We also present the relationship between computing time and drone number in the swarm in circular formation based on 100 repeated simulations in Tab.\ref{table:computing-time}. The GAP could be regarded as real-time when the drone number is small (under 8) while optimization time is too long for onboard deployment for a larger drone number. 

\begin{table}[h]
\caption{Relationship Between Computing Time and Agent Number}
\label{table:computing-time}
\begin{center}
\begin{tabular}{c|c|c|c|c}
\hline
 Agent number & 4  & 6  & 8  & 10  \\
\hline
 Mean computing time (s) & 0.0004  & 0.0010  & 0.0240  & 1.6553  \\
\hline
 \makecell[c]{STD of computing time (s) }& 0.0002 & 0.0009 & 0.0025 & 0.1295 \\
\hline
\end{tabular}
\end{center}
\end{table}

The experiments are conducted with four drones. Before takeoff, each drone will compute the relative positions of the other drones in an egocentric manner, which means they regard their position as the origin. Only horizontal $x$ and $y$ axes are considered because all drones' heights are zero before takeoff. After initialization, each drone can fuse VIO measurements and provide real-time relative positions of the other drones. Then the aerial swarm will take off to perform formation control tasks. In all experiments, relative localization's average onboard optimization time is under 3 ms, which could be regarded as real-time.

\begin{figure}[htpb]
      \centering
      \subfigure[Side view]
        {\includegraphics[scale=0.19]{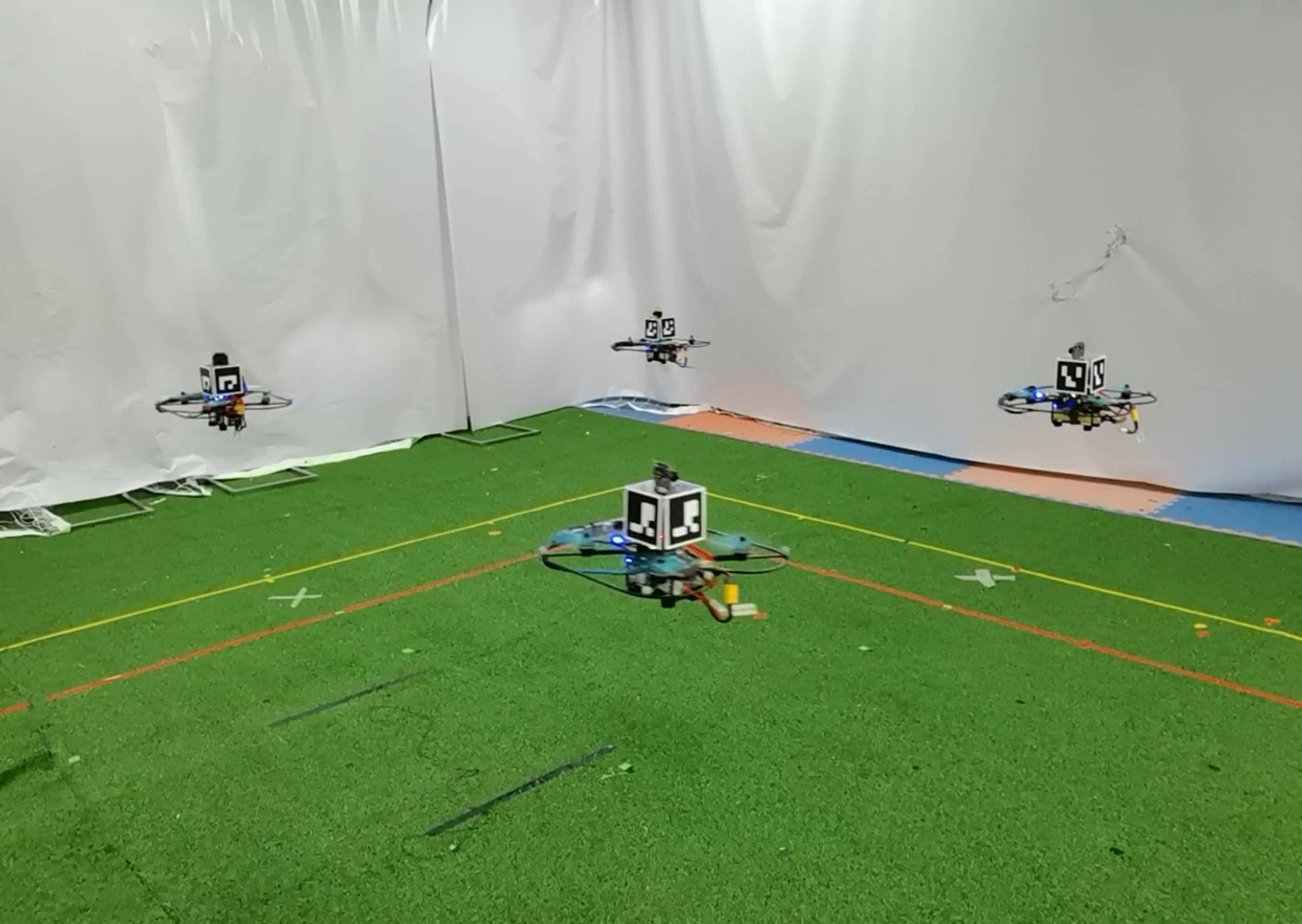}}
      \subfigure[Top view]
        {\includegraphics[scale=0.17]{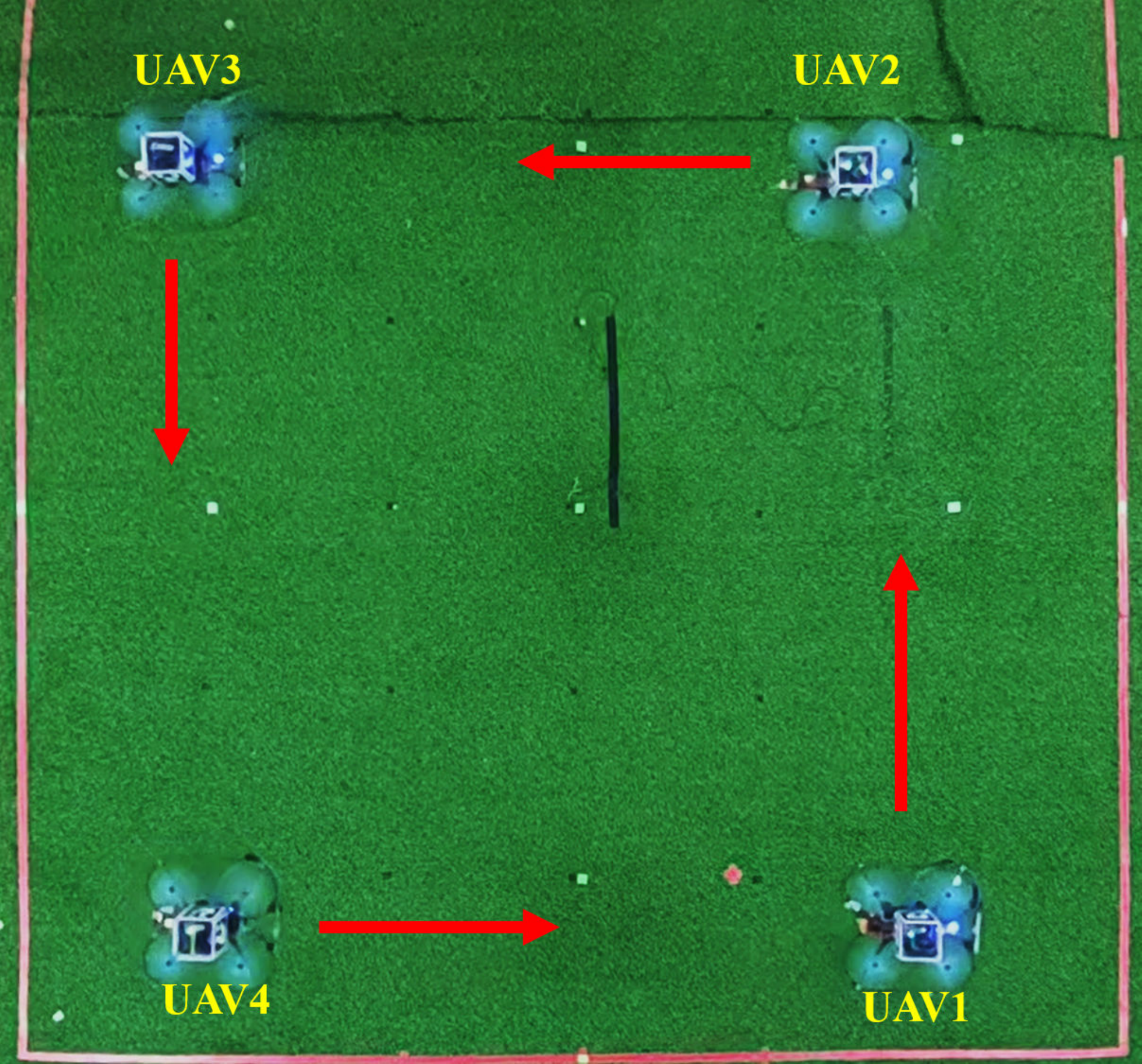}}
      \caption{Four drones in the aerial swarm performing indoor formation control task. Each drone observes the direction of the next drone according to the optimal result of the graph-based attention planning.}
      \label{fig:swarm-cut}
\end{figure}

\begin{figure}[htbp]
      \centering
      \subfigure[Drone 1 view]
        {\includegraphics[scale=0.095]{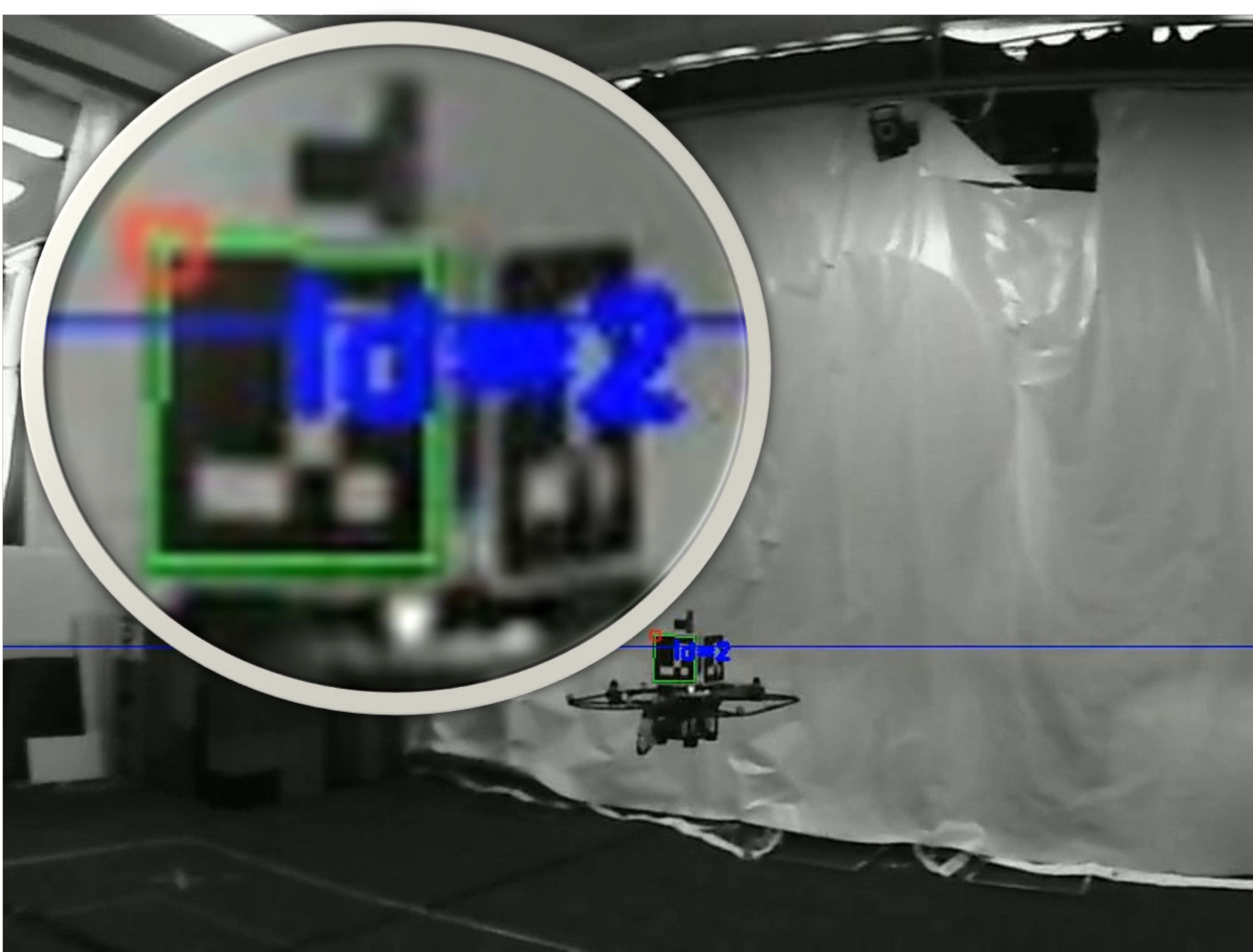}}
      \subfigure[Drone 2 view]
        {\includegraphics[scale=0.095]{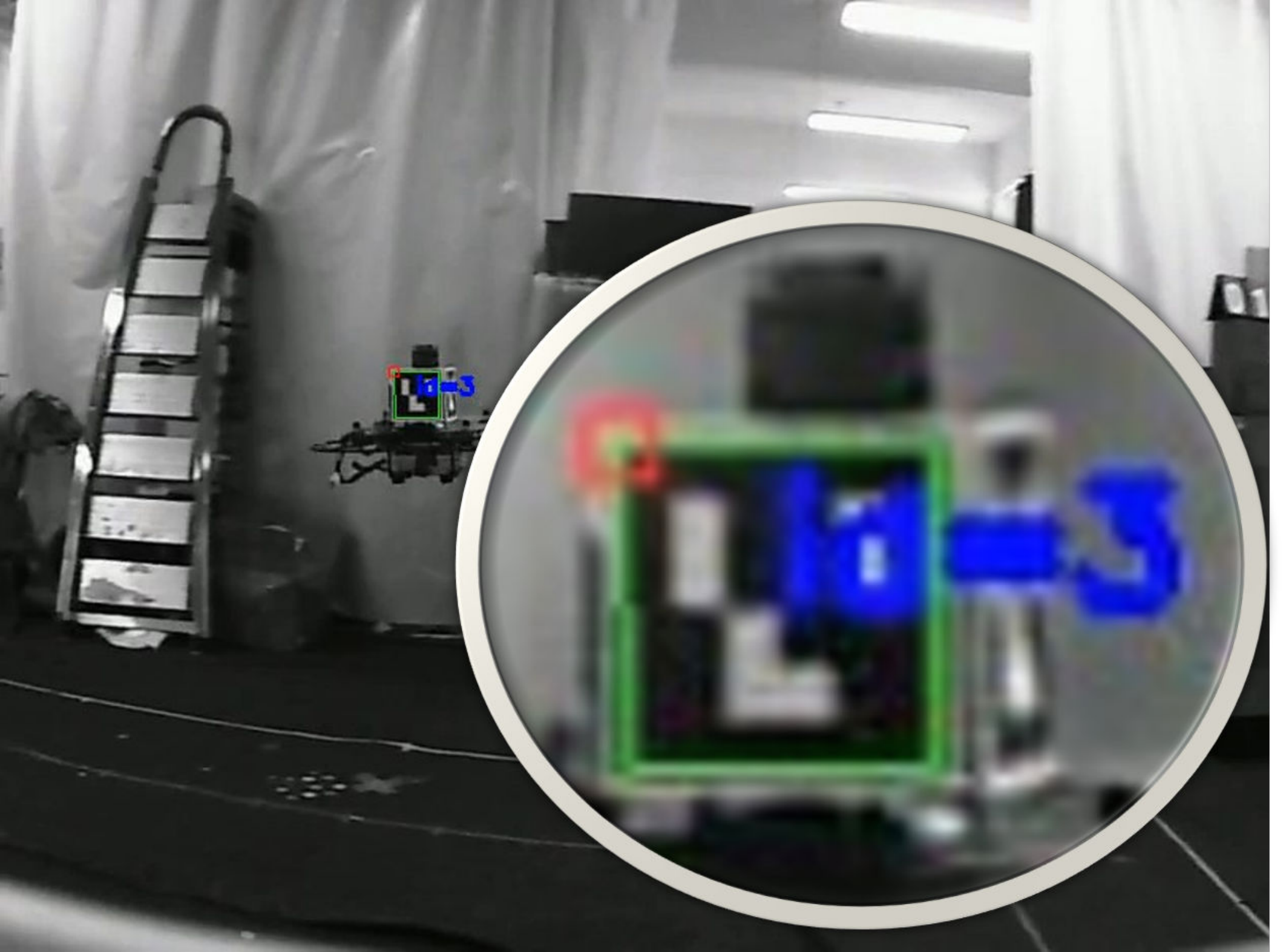}}
      \subfigure[Drone 3 view]
        {\includegraphics[scale=0.095]{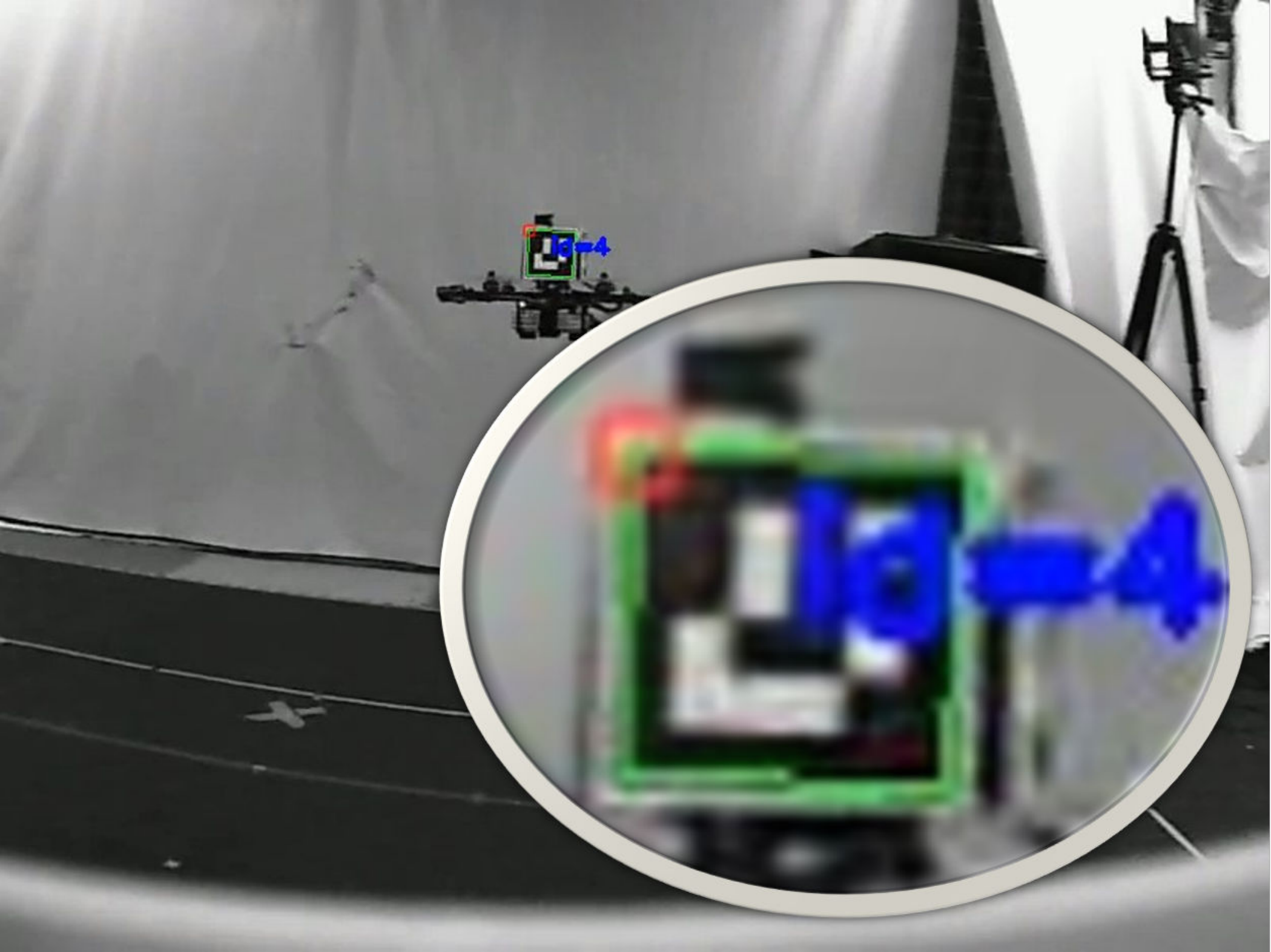}}
    \subfigure[Drone 4 view]
        {\includegraphics[scale=0.095]{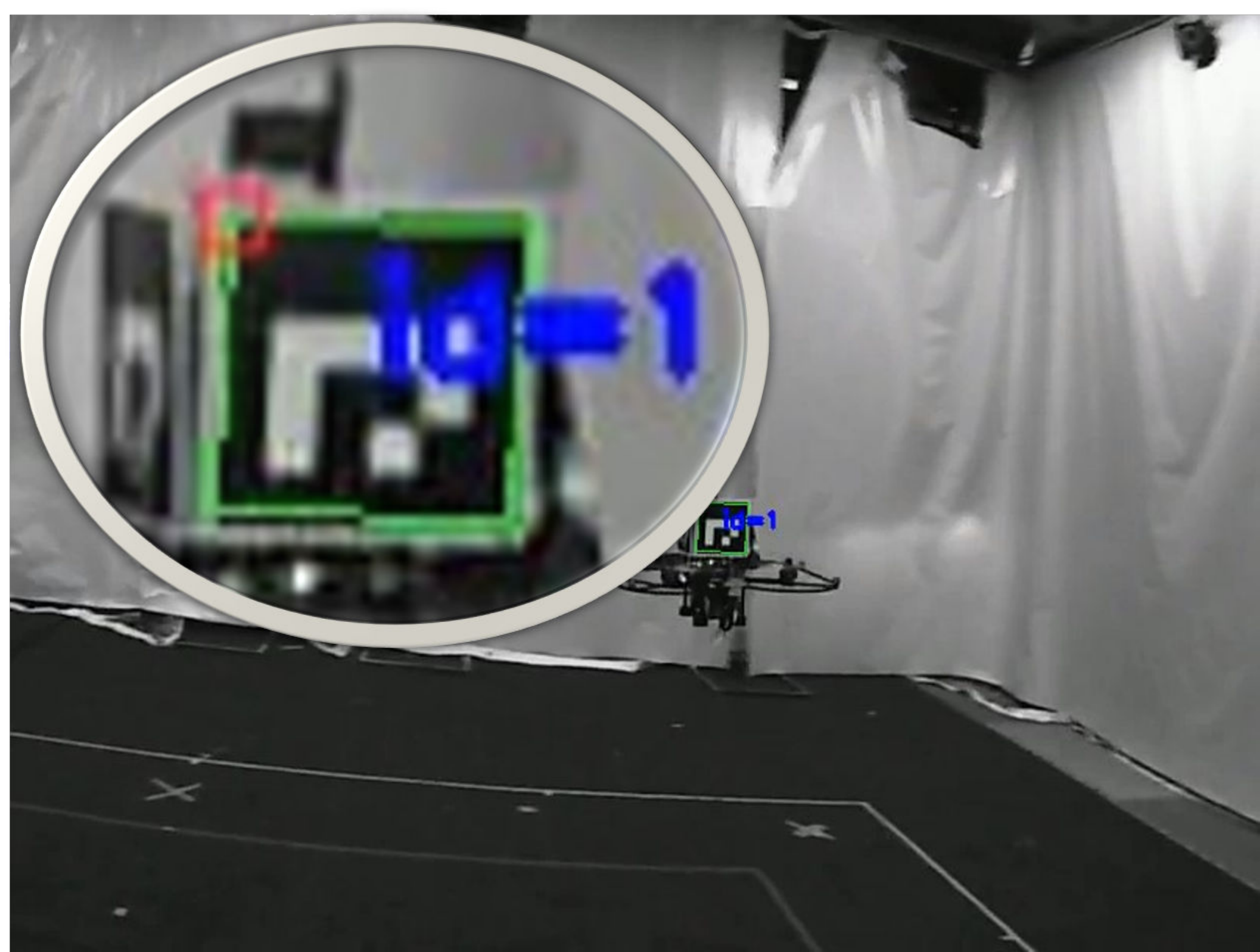}}
      
      \caption{The first person view (FPV) of four drones in the aerial swarm. The ID number and marker position are emphasized in the figure.}
      \label{fig:four}
\end{figure}

We compare the performance of the proposed active vision with fixed vision in the circular formation. The radius of the circle is 1 m. Each drone will accelerate to desired speed from static and maintain the formation at that speed. According to the simulation of GAP, the optimal observation scheme in the formation control task is to detect the direction of the next drone. Fig.\ref{fig:swarm-cut} shows an indoor circular formation from the top view and side view and Fig.\ref{fig:four} shows the first-person view (FPV) of the four drones, respectively. 

\begin{figure}[htbp]
    \centering
    \subfigure[Relative position estimations and errors of active vision system.]
    {\includegraphics[scale=1.2]{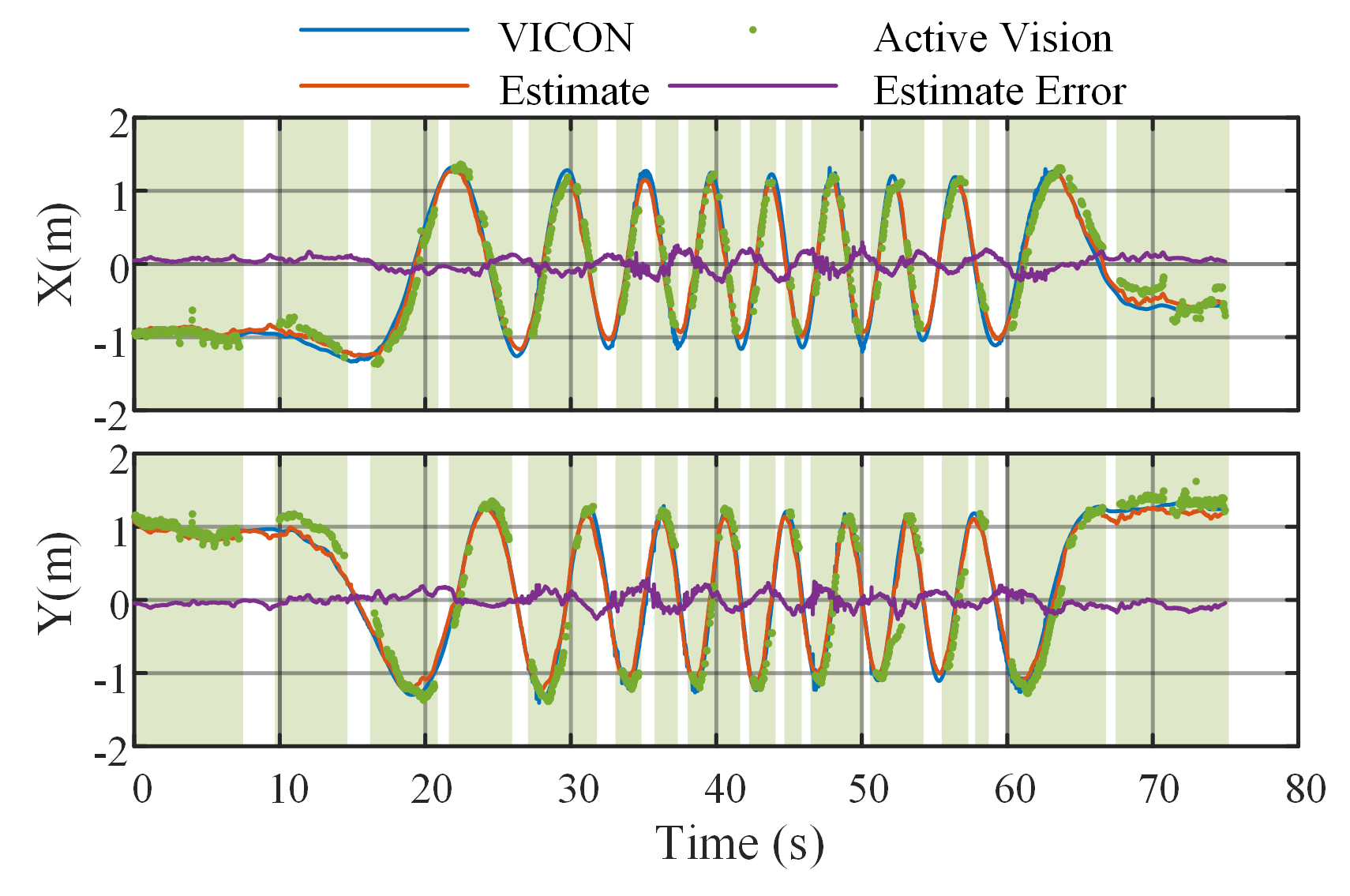}}
    \subfigure[Relative position estimations and errors of fixed vision system.]
    {\includegraphics[scale=1.2]{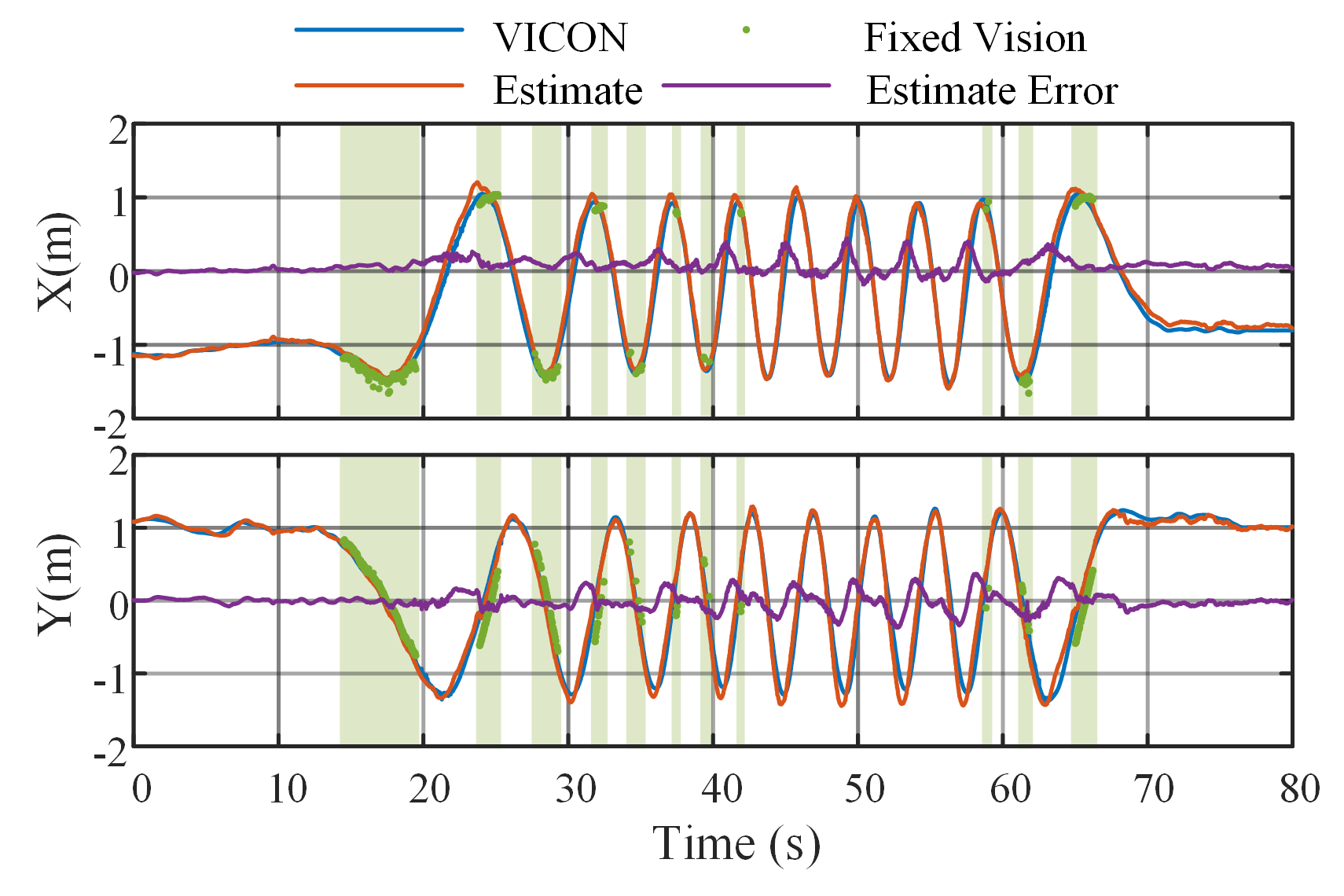}}
%     \subfigure[Time evolution of angles between agents with active vision-based system. ]
%     {\includegraphics[scale=0.95]{angle-active.pdf}}
%     \subfigure[Time evolution of angles between agents with fixed camera system. ]
%     {\includegraphics[scale=0.95]{angle-fix.pdf}}
    
    \caption{The comparison of the proposed active vision-based relative localization and fixed camera system between drone $N_1$ and $N_2$ in the formation flight with 4 drones. Green dots mark the visual detection result, and the green shade represents the time interval of detection.}
    \label{fig:compare}
\end{figure}

Fig.\ref{fig:compare} shows relative localization accuracy of both active vision and fixed vision at 2 m/s speed in one experiment. The estimated trajectory is between drone $N_1$ and $N_2$. Compared with fixed vision, the active vision system outperforms in terms of estimation accuracy and duration of visual detection (marked with green shade). The active vision system has less invisible time interval, marked by more detection data. Also, the RMSE of the active vision system of $x$ and $y$ is 0.096 m and 0.088 m, respectively, while the fixed camera system has 0.121 m and 0.114 m RMSE.

\begin{figure}[htbp]
      \centering
      \includegraphics[scale=1]{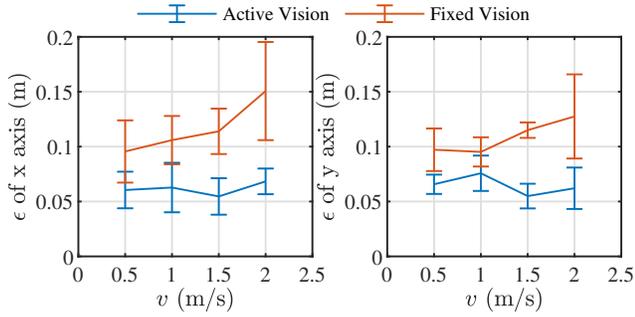}
      \caption{Position estimation error of active and fixed vision at different velocities.}
      \label{fig:est-err-cmp}
\end{figure}

To demonstrate the accuracy of the relative position estimation of active vision, we further compared the performance of active vision and fixed vision under four different velocity conditions, and the results are shown in Fig.\ref{fig:est-err-cmp}. We did 6 replicate trials for each parameter setting. The results show that the estimation errors of active vision are smaller than those of fixed vision under different velocity conditions. As the velocity increases, the estimation error of fixed vision increases, while the estimation accuracy of active vision is not affected. The deterioration of fixed vision estimation could be due to the drift of VIO with increasing velocity. The active vision could compensate for this drift by the active vision detection. For higher velocity, we also did experiments at the velocity of 2.5 m/s and a radius of 1 m. The motion of the UAV was so aggressive that the UAV's inclination was large (over 40$^{\circ}$) and the active vision system was not able to observe other drones effectively. This led to formation divergence with VIO's drifting away.

\begin{figure}[htbp]
      \centering
      \includegraphics[scale=1]{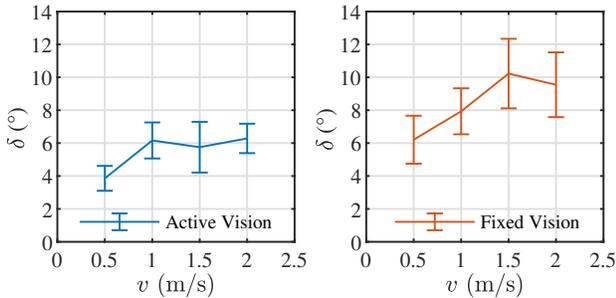}
      \caption{Formation angle error of active and fixed vision at different velocities.}
      \label{fig:angle-err-box}
\end{figure}

\begin{figure}[htbp]
      \centering
      \includegraphics[scale=1]{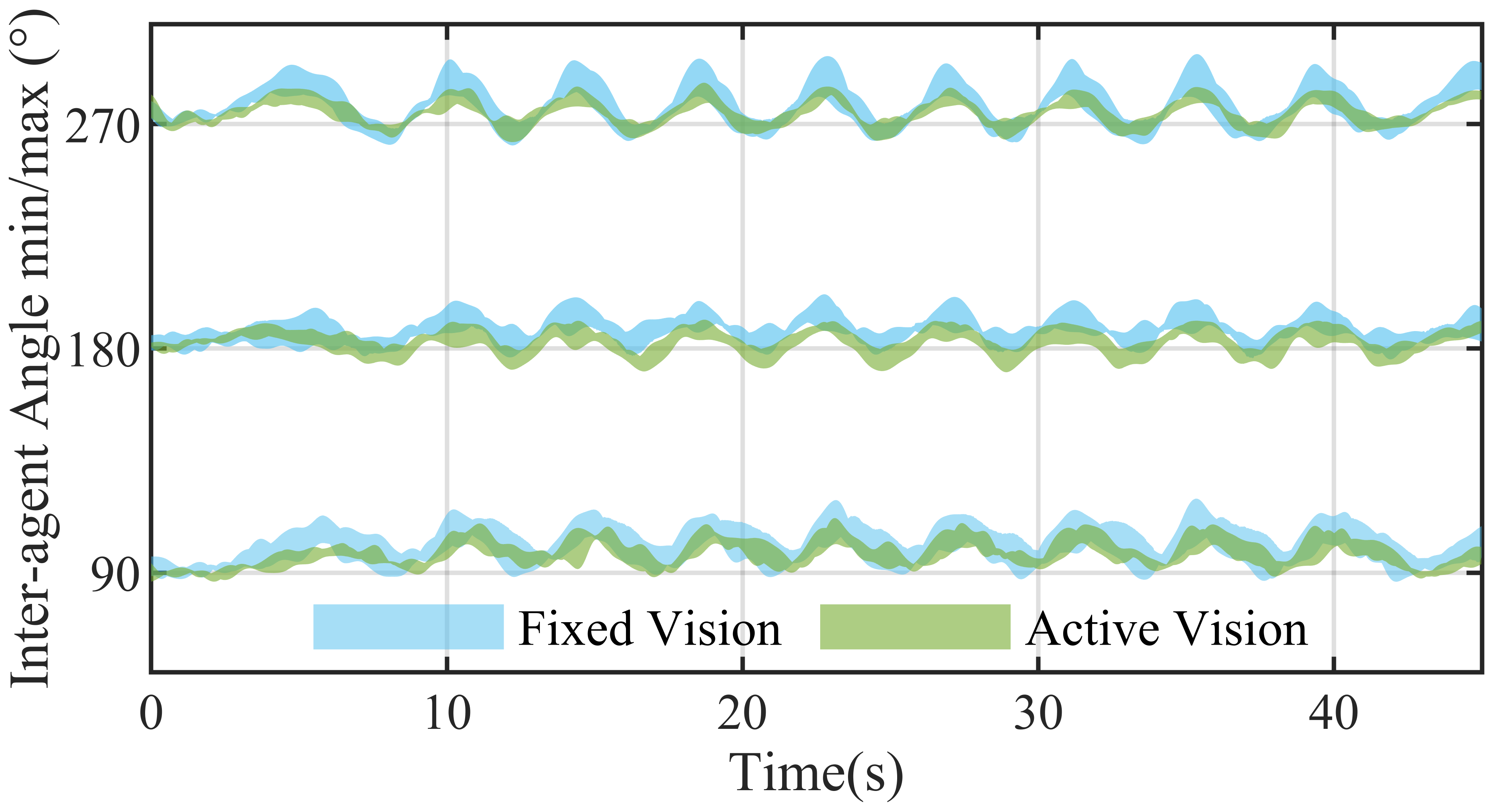}
      \caption{Inter-agent angle range of active vision and fixed vision in 6 trials of circular formation at 1.5 m/s. Agent 1 is the reference and the formation angles of the other three drones should be 90$^{\circ}$, 180$^{\circ}$, and 270$^{\circ}$}
      \label{fig:angleminmax}
\end{figure}

In addition, we compared the formation accuracy of active vision and fixed vision under four different velocity conditions in Fig.\ref{fig:angle-err-box}. Angles between each drone represent the formation accuracy. In the circular formation of four drones, the angle between neighbors is supposed to be 90$^{\circ}$. As the velocity increases, the error of the active vision formation is less affected, while the error of the fixed vision increases. Due to feedback in the formation control law, the active vision system has more observations during experiments and is thus more accurate. In particular, we show the curves of formation angles of active vision and fixed vision at the velocity of 1.5 m/s in repeated experiments in Fig.\ref{fig:angleminmax}. Using drone $N_1$ as a reference, the formation angles of the other three drones should be 90$^{\circ}$, 180$^{\circ}$, and 270$^{\circ}$, respectively. In Fig.\ref{fig:angleminmax}, the formation of active vision is closer to target values and has less fluctuation and lower variance.

% \begin{figure*}[htp]
%     \centering
%     \subfigbottomskip=8pt
%     \subfigure[Trajectory of four agents on xy plane.]
%     {\includegraphics[scale=0.55]{Pxy-plane.pdf}
%     \label{subfig:Pxy-plane}
%     }
%     \quad
%     \subfigure[Time evolution of velocity of four agents.]
%     {\includegraphics[scale=0.55]{Vxytplane.pdf}
%     \label{subfig:Vxytplane}
%     }
%     \quad
%     \subfigure[Velocity curve of four agents on xy plane.]
%     {\includegraphics[scale=0.55]{Vxyplane.pdf}
%     \label{subfig:Vxyplane}
%     }
%     \caption{The trajectory and velocity curve during the formation flight of four agents.}
%     \label{fig:Pxy and Vxy}
% \end{figure*}

% The intialization of relative positions is tested with 4 drones before takeoff. Fig.TODO shows the estimated intial positions and those recorded by the motion capture system. The error comparison between the proposed method and the ground truth shows that that the RMSE can reach TODOcm level for each agent, which lay a solid basis for subsequent relative localization.

We also provide a detailed error comparison between the active vision, fixed vision and the ground truth in Tab.\ref{tab:cmp}. The UWB data and the active vision detection data are abandoned in different scenarios to verify the effectiveness of the proposed sensor fusion method and the contribution of different measurements. According to Tab.\ref{tab:cmp}, it is shown that the RMSE of the active vision can reach 5 cm level at different velocities. The absence of UWB slightly affects the estimation accuracy, which may be due to the large variance of the distance estimation of the UWB hardware used in experiments. Meanwhile, the absence of active vision causes an obvious deterioration of the estimation accuracy with an error greater than 10 cm.

\begin{table}
\centering
\caption{Comparison of Relative Position Estimation with Groundtruth at Different Velocities}
\label{tab:cmp}
\begin{tabular}{c|cc|ccc|c} 
\hline
\begin{tabular}[c]{@{}c@{}}Vel\\(m/s)\end{tabular} & \multicolumn{2}{c|}{\begin{tabular}[c]{@{}c@{}}Evaluation\\Metrics \\(m)\end{tabular}} & \begin{tabular}[c]{@{}c@{}}\textbf{Proposed}\\\textbf{Method}\end{tabular} & \begin{tabular}[c]{@{}c@{}}Without\\UWB\end{tabular} & \begin{tabular}[c]{@{}c@{}}Without\\Active \\Vision\end{tabular} & \begin{tabular}[c]{@{}c@{}}Fixed\\Vision\end{tabular}  \\ 
\hline
\multirow{4}{*}{0.5}                               & \multirow{2}{*}{x} & RMSE                                                              & \textbf{0.060}                                                             & 0.066                                                & 0.136                                                            & 0.096                                                  \\
                                                   &                    & STD                                                               & 0.017                                                                      & 0.017                                                & 0.020                                                            & 0.028                                                  \\ 
\cline{2-7}
                                                   & \multirow{2}{*}{y} & RMSE                                                              & \textbf{0.066}                                                             & 0.083                                                & 0.132                                                            & 0.097                                                  \\
                                                   &                    & STD                                                               & 0.009                                                                      & 0.018                                                & 0.022                                                            & 0.019                                                  \\ 
\hline
\multirow{4}{*}{1}                                 & \multirow{2}{*}{x} & RMSE                                                              & \textbf{0.059 }                                                            & 0.083                                                & 0.133                                                            & 0.106                                                  \\
                                                   &                    & STD                                                               & 0.016                                                                      & 0.013                                                & 0.019                                                            & 0.022                                                  \\ 
\cline{2-7}
                                                   & \multirow{2}{*}{y} & RMSE                                                              & \textbf{0.077}                                                             & 0.091                                                & 0.146                                                            & 0.095                                                  \\
                                                   &                    & STD                                                               & 0.018                                                                      & 0.013                                                & 0.015                                                            & 0.013                                                  \\ 
\hline
\multirow{4}{*}{1.5}                               & \multirow{2}{*}{x} & RMSE                                                              & \textbf{0.055 }                                                            & 0.057                                                & 0.120                                                            & 0.114                                                  \\
                                                   &                    & STD                                                               & 0.017                                                                      & 0.016                                                & 0.019                                                            & 0.021                                                  \\ 
\cline{2-7}
                                                   & \multirow{2}{*}{y} & RMSE                                                              & \textbf{0.055}                                                             & 0.060                                                & 0.115                                                            & 0.115                                                  \\
                                                   &                    & STD                                                               & 0.011                                                                      & 0.013                                                & 0.017                                                            & 0.007                                                  \\ 
\hline
\multirow{4}{*}{2}                                 & \multirow{2}{*}{x} & RMSE                                                              & \textbf{0.071}                                                             & 0.082                                                & 0.104                                                            & 0.151                                                  \\
                                                   &                    & STD                                                               & 0.012                                                                      & 0.018                                                & 0.020                                                            & 0.045                                                  \\ 
\cline{2-7}
                                                   & \multirow{2}{*}{y} & RMSE                                                              & \textbf{0.056}                                                             & 0.070                                                & 0.092                                                            & 0.127                                                  \\
                                                   &                    & STD                                                               & 0.007                                                                      & 0.008                                                & 0.014                                                            & 0.038                                                  \\
\hline
\end{tabular}
\end{table}

% Finally, we show the result of formation control by providing trajectory and velocity data in Fig.\ref{fig:Pxy and Vxy}. Fig.\ref{subfig:Pxy-plane} is the trajectory of four agents in XY plane. Fig.\ref{subfig:Vxytplane} is the velocity of $x$ and $y$ axis with time resolution. The figure is spindle-like because four agents accelerate from a static state and end in the static state. Fig.\ref{subfig:Vxyplane} shows the velocity curve of four agents on $xy$ plane. The maximum acceleration is 4 $m/s^2$. The precision and consistency demonstrated above show the potential of applying the active vision system in the high-speed formation control task.

\section{Conclusion}
This paper proposes a novel active vision-based relative localization framework to tackle the restricted FOV of the vision-based approaches. We devise GAP to obtain the optimal active vision planning of the swarm. Measurements from active vision, VIO and UWB are fused to obtain relative positions, achieving 5 cm RMSE at different velocities. The result of relative localization is integrated with the formation control task to perform agile formation. Simulations and experiments validate the effectiveness of the proposed active vision system compared with the fixed vision system in terms of detection duration, estimation and formation accuracy at different velocities. 

In the future, we plan to improve the computing efficiency of the GAP further so that it is scalable to large-scale swarms. Also, we would apply our active vision system to aggressive motion with higher speed by addressing the vision loss in this scenario.

\addtolength{\textheight}{-6.5cm}   % This command serves to balance the column lengths

%%%%%%%%%%%%%%%%%%%%%%%%%%%%%%%%%%%%%%%%%%%%%%%%%%%%%%%%%%%%%%%%%%%%%%%%%%%%%%%%

%%%%%%%%%%%%%%%%%%%%%%%%%%%%%%%%%%%%%%%%%%%%%%%%%%%%%%%%%%%%%%%%%%%%%%%%%%%%%%%%


% \section*{Acknowledgment}
% We thank Ziwei Zhou for the help with conducting indoor experiments and video capture, as well as Sensen Liu, Zhaoying Wang, Ziying Lin for the helpful discussions.

\bibliographystyle{IEEEtran}
\bibliography{IEEEabrv,mybib}

\end{document}